\documentclass[letterpaper, 10 pt, conference]{ieeeconf} 
\usepackage[T1]{fontenc}
\usepackage{amsmath,amsfonts}
\usepackage{array}
\usepackage[caption=false,font=normalsize,labelfont=sf,textfont=sf]{subfig}
\usepackage{textcomp}
\usepackage{stfloats}
\usepackage{url}
\usepackage{verbatim}
\usepackage{graphicx}
\usepackage{amsmath}
\usepackage{bm}
\usepackage[ruled,linesnumbered]{algorithm2e}
\usepackage{multirow}
\usepackage{hyperref}
\usepackage{cleveref}

\def\BibTeX{{\rm B\kern-.05em{\sc i\kern-.025em b}\kern-.08em
    T\kern-.1667em\lower.7ex\hbox{E}\kern-.125emX}}
\usepackage{balance}

\IEEEoverridecommandlockouts
\title{\LARGE \bf
Application of LLM Guided Reinforcement Learning in Formation Control with Collision Avoidance
}

\author{Chenhao Yao, Zike Yuan, Xiaoxu Liu$^*$, Chi Zhu
\thanks{This work was supported by National Natural Science Foundation of China under Grant 62003218, Stable Support Projects for Shenzhen Higher Education Institutions under Grant 20220717223051001.}
\thanks{The authors are with the School of Sino-German College of Intelligent Manufacturing, Shenzhen Technology University, Shenzhen 518118, China. Emails: yaochenhao@sztu.edu.cn; yuanzike2022@email.szu.edu.cn; liuxiaoxu@sztu.edu.cn; zhuchi@sztu.edu.cn.}
}

\begin{document}

\maketitle 
\thispagestyle{empty} 
\pagestyle{empty} 

\begin{abstract}
Multi-Agent Systems (MAS) excel at accomplishing complex objectives through the collaborative efforts of individual agents. Among the methodologies employed in MAS, Multi-Agent Reinforcement Learning (MARL) stands out as one of the most efficacious algorithms. However, when confronted with the complex objective of Formation Control with Collision Avoidance (FCCA): designing an effective reward function that facilitates swift convergence of the policy network to an optimal solution.
In this paper, we introduce a novel framework that aims to overcome this challenge. By giving large language models (LLMs) on the prioritization of tasks and the observable information available to each agent, our framework generates reward functions that can be dynamically adjusted online based on evaluation outcomes by employing more advanced evaluation metrics rather than the rewards themselves. This mechanism enables the MAS to simultaneously achieve formation control and obstacle avoidance in dynamic environments with enhanced efficiency, requiring fewer iterations to reach superior performance levels. Our empirical studies, conducted in both simulation and real-world settings, validate the practicality and effectiveness of our proposed approach.
\end{abstract}
\textbf{Project Website}: \href{https://macsclab.github.io/LLM_FCCA}{https://macsclab.github.io/LLM\_FCCA}

\section{INTRODUCTION}
Multi-Agent Systems (MAS) have demonstrated superior performance across various fields, 
showcasing higher task efficiency and stronger fault tolerance compared to single-agent systems. However, current MAS applications predominantly operate within structured environments, with less satisfactory performance in more complex, unstructured scenarios. Traditional methods, such as Optimal Reciprocal Collision Avoidance (ORCA) \cite{ORCA} and Artificial Potential Fields (APF) \cite{APF} , explicitly model the system and precisely calculate the instructions for each agent at every moment. However, these approaches often rely on certain assumptions (e.g., ORCA assumes that all agents follow the same obstacle avoidance strategy), which can lead to policy failures when real-world deployments do not align with these preconditions.

In recent years, Multi-Agent Reinforcement Learning (MARL), a type of reinforcement learning (RL) has made notable advancements in addressing challenges such as formation control, obstacle avoidance, and maintaining system stability within MAS, showcasing its substantial potential. Through continuous interaction with their environment, agents iteratively refine their policies to maximize cumulative rewards, highlighting the significant impact that the design of reward function can have on policy quality. While crafting an effective reward function is relatively straightforward for single-task scenarios, the Formation Control with Collision Avoidance (FCCA) problem introduces the complexity of accounting for interdependencies between different tasks. This added layer of complexity often necessitates considerable time and effort in designing and tuning reward functions; even minor adjustments, such as tweaking the weights of individual reward components, can require retraining of the models based on the original models to ensure optimal performance across multiple objectives.

\begin{figure*}[!t]
    \centering
    \includegraphics[width=18cm]{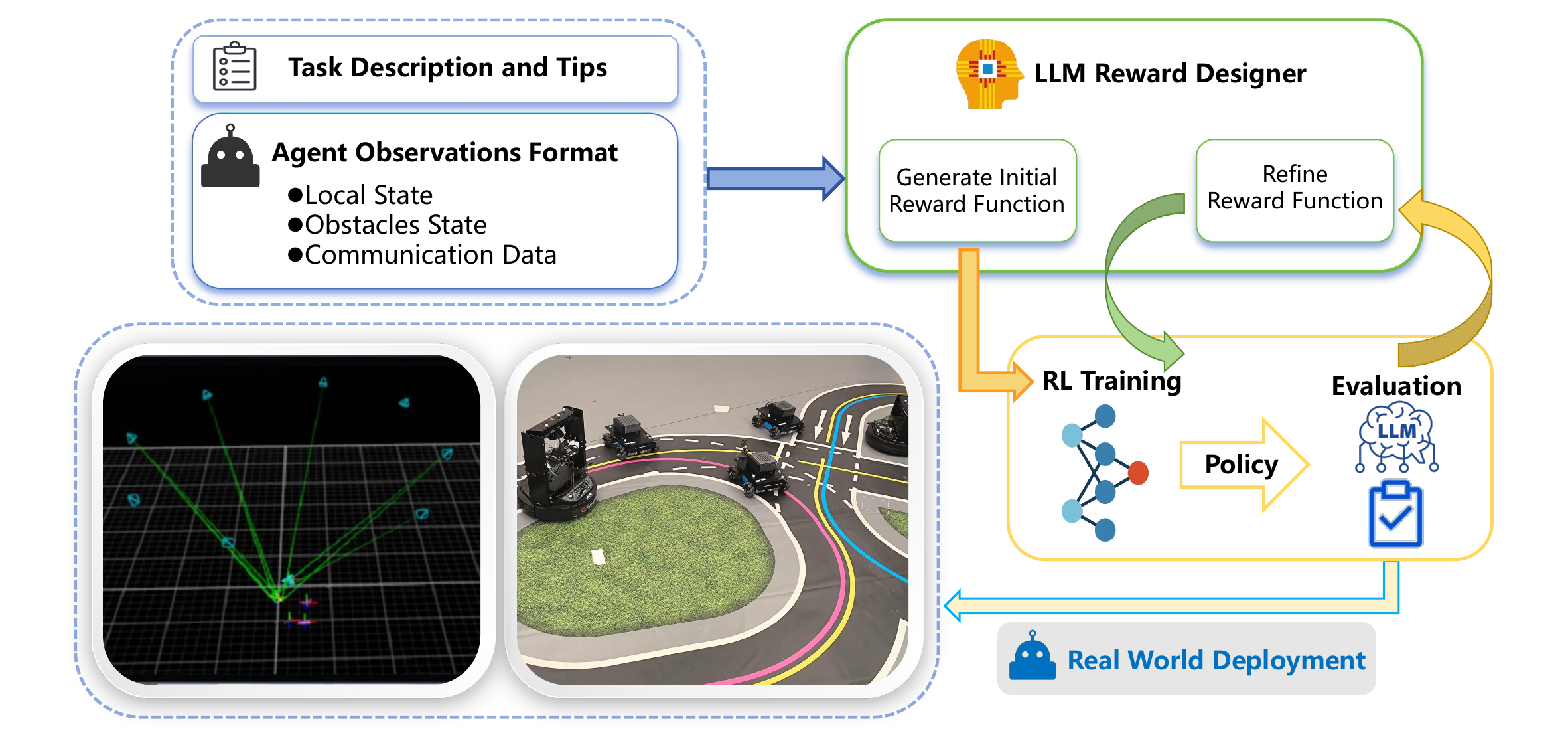}
    \caption{Overview of our method}
    \label{overview}
\end{figure*}

To address the above issues, we propose a framework where agent observations are provided to an LLM, which generates an initial reward function focused on core objectives rather than optimal performance across all tasks. Instead of relying on reward magnitude, effectiveness is evaluated by how well predefined performance criteria are met. After a fixed number of iterations, these criteria and task-specific rewards are fed back to the LLM, enabling online adjustments to improve the reward function. We validate our method in a complex scenario where a MAS maintains formation, avoids dynamic obstacles, and reaches its destination in minimal time with stable actions. An overview of the proposed method is shown in Fig.~\ref{overview}.

In summary, the contributions of this work can be highlighted as follows:
\begin{itemize} 
\item We are the first to apply LLM-guided RL to the multi-agent FCCA task, enabling the creation of sophisticated reward structures that guide agents in achieving complex objectives.
\item We implemented a framework that dynamically updates reward functions, allowing continuous improvement and higher efficiency with fewer iterations.
\item We validated the effectiveness and practicality of our approach through the use of sim-to-sim and sim-to-real validation methods.
\end{itemize}


\section{RELATED WORKS}
\label{RELATED_WORKS}

\subsection{Multi-Agent Reinforcement Learning}

With the proposal and refinement of various effective RL algorithms, such as Proximal Policy Optimization (PPO) \cite{PPO} , Importance Weighted Actor-Learner Architecture(IMPALA) \cite{impala}, and QMIX \cite{QMIX}, MARL has made significant progress. These approaches employ neural networks to approximate the policy of an agent rather than modeling the entire task as an optimization problem.

Initially, MARL was primarily applied in video games, like StarCraft II \cite{StarCraft} and DOTA II \cite{DOTA2}. Furthermore, in \cite{mappo_games} , the researchers conducted a series of extensive experiments utilizing the Proximal Policy Optimization (PPO) algorithm across a multitude of distinct cooperative multi-agent benchmarks. The experimental results consistently demonstrated superior performance, with the level of coordination among the agents even exceeding that of human teams. However, these achievements were not substantiated by real-world deployments, leaving questions about the efficacy of these RL algorithms in real-world robotics.

Utilizing a hybrid approach that integrates imitation learning (IL) and RL, the researchers \cite{ORCA-RL} trained MAS for tasks FCCA, resulting in markedly improved performance in maintaining formations and avoiding collisions compared to earlier methods, while also demonstrating that models trained with MARL methodologies possess robust generalization capabilities and achieved satisfactory performance in real-world experiments. However, the effectiveness of these systems remains constrained by the leader-follower architecture, as the system can become unstable if the leader makes incorrect decisions or fails to effectively transmit decision information to the followers, potentially leading to a loss of control over the entire system.

By adopting a distributed architecture in \cite{mappo_rw}, the researchers facilitated independent decision-making for each agent, thus mitigating the risk of system instability due to a leader's failure. Curriculum learning was utilized to streamline the training process, incrementally raising the complexity of objectives faced by agents. However, the real-world experimental scenarios were relatively simplistic, featuring only sparse static obstacles, which may not sufficiently demonstrate the superiority of MARL approaches. Moreover, individual reward components converged more slowly at the beginning due to the necessity of learning multiple objectives concurrently, raising concerns that in more complex environments, the system might become trapped in local optima, thereby reducing success rates.
In the context of tackling complex objectives, it becomes clear that the design of reward functions is one of the most crucial elements in MARL.

\subsection{LLM for Robotics}

The intersection of LLM with various disciplines represents an emerging and popular frontier in research; at present, LLM have already acquired substantial semantic knowledge and exhibit formidable text generation capabilities. Within the framework described \cite{google_saycan}, LLM have been employed to generate a series of high-level action sequences as instructions, and to pre-train a suite of low-level skills through RL. Through these high-level action sequences, LLM guide agents in selecting the appropriate low-level skills to perform corresponding natural language actions, thereby achieving real-world grounding. Based on this, the CodeAct method \cite{CodeAct} allows for online adjustment of actions based on observed information within the framework of SayCan. Meanwhile, \cite{LLM_nav} breaks down a navigation command into multiple sub-commands, enabling the intelligent agent to learn navigation strategies that adapt to environmental changes. However, these methodologies necessitate additional learning for low-level actions and cannot be directly deployed on agents.

On the other hand, \cite{text2reward} and \cite{eureka} directly utilizes the LLM to generate the reward functions required for RL, and employs these reward functions to learn low-level actions. The results from validation across several challenging tasks indicate that models trained using this method achieve performance that is comparable to or surpasses that of human experts in executing complex objectives. However, generating reward functions in a one-time process for training and simply feeding back the magnitude of rewards obtained from LLM-generated reward functions after environmental evaluation to the LLM like the he aforementioned works lead to tendency to get trapped in local optima when dealing with complex objectives, i.e., the agents focus on completing only a single task while neglecting others.

In extracting the advantages of the aforementioned approaches, we have established high-level evaluation metrics for each task and permitted the LLM-generated reward functions to initially focus on critical or more challenging-to-converge tasks. Following this initial phase, the reward functions are progressively fine-tuned to address secondary objectives. This method ultimately enables superior performance across all tasks.

\section{PRELIMINARIES}
\label{PRELIMINARIES}

\subsection{Formation Control with Collision Avoidance}

FCCA is a classic multi-agent coordination problem with applications spanning ground-based robots\cite{fcca_car}, aerial drones\cite{racer}, and surface vehicles\cite{fcca_usv}. In the absence of obstacles, multiple agents need to obtain positional information from each other to maintain a good formation and collectively move toward the goal. When obstacles appear, one or more agents must temporarily disrupt the formation to avoid the obstacles, creating a conflict between “maintaining formation” and “avoiding obstacles.” After leaving the obstacle area, there is also a conflict between “reaching the goal” and “maintaining formation.” For static obstacles, the issue can be explicitly formulated as an optimization problem through pre-planned paths; however, for dynamic obstacles, especially in unstructured environments unlike roadways, real-time path planning becomes significantly challenging. To address the FCCA problem with dynamic obstacles, we have opted for an RL-based approach.

\subsection{Reinforcement Learning Modeling}
RL is typically formalized theoretically as a Markov Decision Process (MDP) \cite{MDP}: $(\mathcal{S}, \mathcal{A}, P, R, \mathcal{O})$, which means that each agent selects actions $\mathcal{A}$ based on its own observations $\mathcal{O}$ to maximize the reward $R$.The other variables $(\mathcal{S}$ represent the global state, which include the observations of all agents, and $P$ is the state transition function. For each agent $i$, our objective is to learn an individual policy $\pi_{\theta_i}(\bm{a_t} | \bm{o_t})$, where $\theta_i$ are the network parameters of the corresponding agent's policy network such that the agent can take actions $\bm{a_t}$ based on its observations $\bm{o_t}$ at the time step $t$. Similarly, the value function can be parameterized as: $V_\omega^i(\mathcal{S})$.

\section{METHODOLOGY}
\label{METHODOLOGY}

\subsection{Agents Observations as Context}
\begin{figure}[t]
    \centering
    \includegraphics[width=9cm]{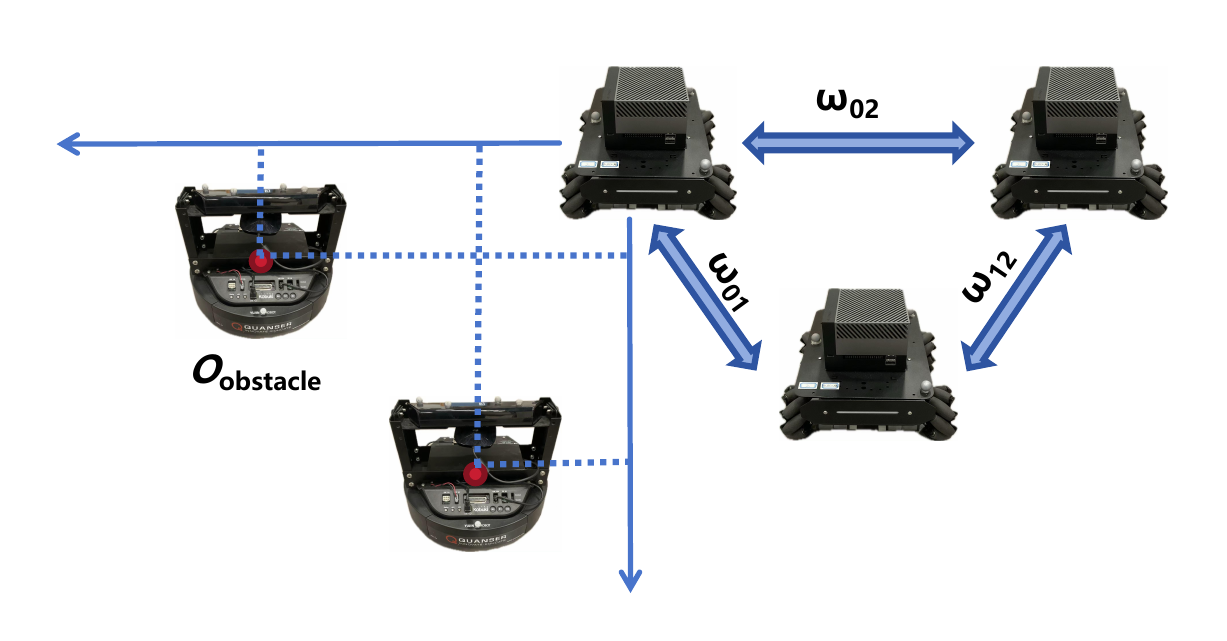}
    \caption{Diagram of the agents observatios}
    \label{observations}
\end{figure}

To enable the LLM to generate executable reward functions, it is necessary to provide context to the LLM. In \cite{text2reward} and \cite{eureka}, the entire environmental information is input into the LLM. However, in distributed MAS, each agent typically gathers its own local observations, observations of obstacles within its vicinity, as well as data acquired through communication with neighboring agents, a diagram is shown in Fig. \ref{observations}. Specifically, for each individual agent, the available own local observations are comprised of: the coordinates of the destination $g_x, g_y$  as well as the velocity $v$ and orientation $\theta$ acquired from the policy network; the observations of obstacles include: the position of the obstacles in the agent's local coordinate system $p^i_{ox}, p^i_{oy}$, the velocity in the global coordinate system $v^i_{ox}, v^i_{oy}$, which are obtained through differencing positions.

The data obtained through communication with neighboring agents consist solely of the coordinates of those adjacent agents. However, designing a reward function directly from the raw coordinate information would involve mathematical derivations related to graph theory. In the absence of human feedback, LLM can understand and make efficient use of this information. Therefore, we preprocess the formation information acquired through communication according to the method outlined in \cite{zju_uav_fcca}.

Given that communication between agents in MAS is bidirectional, the entire system which includes $ N $ agents can be modeled as an undirected graph \( \mathcal{G} = (\mathcal{V}, \mathcal{E}) \), where agents are represented as the set of nodes $ \mathcal{V} := \{1, 2, \ldots, N\} $ and communication links between them as the set of edges $ \mathcal{E} \subseteq \mathcal{V} \times \mathcal{V} $. Each edge $ e_{ij} \in \mathcal{E} $ connecting nodes $ i $ and $ j $ signifies that agents $ i $ and $ j $ have the capability to measure their relative distances.Therefore, the weight of each edge can be defined as:
\begin{equation}
  w_{ij} = \| \mathbf{p}_i - \mathbf{p}_j \|^2, \quad (i, j) \in \mathcal{E},  
\end{equation}
where $\mathbf{p}_i$ is the position vector of the agent i, $\| . \|^2$ is Euclidean norm.
Therefore, the Laplacian matrix can be derived from the weights of the edges:
\begin{equation}
    \mathbf{L} = \mathbf{D} - \mathbf{A}, 
\end{equation}
where $\mathbf{A}$ is the adjacency matrix and $\mathbf{D}$ is the degree matrix. Additionally, to adapt to the complex environments with dynamic obstacles, we permit the MAS to perform scaling and rotation while maintaining formation. Therefore, it is also necessary to symmetrically normalize the Laplacian matrix:
\begin{equation}
\hat{\mathbf{L}} = \mathbf{D}^{-1/2} \mathbf{L} \mathbf{D}^{-1/2} = \mathbf{I} - \mathbf{D}^{-1/2} \mathbf{A} \mathbf{D}^{-1/2},
\end{equation}
Next, the formation characteristics can be obtained as:
\begin{equation}
f = \| \hat{\mathbf{L}} - \hat{\mathbf{L}}_{\text{des}} \|_F^2.
\label{formation_error}
\end{equation}
where $\hat{\mathbf{L}}_{\text{des}}$ is the symmetric normalized Laplacian of the desired formation and $\| . \|_F^2$ is the Frobenius norm. 

Finally, the agent's own observations and the formation information are encapsulated into a class, denoted as: $\bm{o_{agent}}=[g_x, g_y, v, \theta]$; the observations of obstacles are encapsulated into another class, denoted as: $\bm{o^i_{obstacle}} = [p^i_{ox}, p^i_{oy}, v^i_{ox}, v^i_{oy}]$; these classes serve as the context input for the LLM.

\subsection{LLM-Based Reward Function Generation}

The generation and selection of the initial reward function are critical. If the LLM fails to accurately understand the task requirements when generating the initial reward function, it will be challenging to achieve results comparable to those crafted by human experts, even with multiple revisions based on feedback test results in the absence of human intervention. Firstly, it is essential to clearly define for the LLM that its role is that of a reward function designer, with the objective of generating a reward function for MARL. This reward function must address multiple objectives simultaneously: 
\begin{itemize}
\item Avoiding dynamic obstacles within the environment.
\item Maintaining the specified formation shape.
\item Reaching the destination.
\end{itemize}
On the condition that the aforementioned tasks are successfully completed, it is also necessary to ensure, as much as possible, the following requirements:
\begin{itemize}
\item Sustaining a stable velocity to facilitate practical deployment.
\item Completing the mission in the shortest possible time.
\end{itemize}

Previous methods often provide LLMs with expert-designed reward templates, which improves executability but risks inheriting design flaws. LLMs may also struggle to handle complex tasks with multiple interacting objectives, potentially omitting key elements. To address this, we start with a simple reward function focused on the most convergent objective, 'reaching the destination', and progressively incorporate other tasks after validating convergence in a basic environment. The overall process is outlined in Algorithm 1.


\begin{algorithm}
\caption{LLM-Guided Reward Initialization}
\label{alg:llm-init}
\KwIn{Observation structure $\mathcal{O}$, task list $\mathcal{T}$, prompt $\mathcal{P}_0$, environment $\mathcal{E}$, threshold $\eta$}
\KwOut{Initial reward $R_0$, policy $\pi_{\theta_i}$, value $\omega_i$}

Initialize $\mathcal{P}_0$ with $\mathcal{O}$ and $\mathcal{T}$\;
\For{$k = 0, 1, 2, \dots$}{
  $R_k \gets \texttt{LLM}(\mathcal{P}_k)$\;
  Train $\pi_{\theta_i}^{(k)}, \omega_i^{(k)}$ in $\mathcal{E}$ using $R_k$\;
  Evaluate success rate $s_k$ and feedback metrics $\mathcal{M}_k$\;
  \If{$s_k \geq \eta$}{
    \textbf{break and return} $R_0 \gets R_k$, $\pi_{\theta_i} \gets \pi_{\theta_i}^{(k)}$, $\omega_i \gets \omega_i^{(k)}$\;
  }
  Update $\mathcal{P}_{k+1}$ using $\mathcal{M}_k$\;
}
\end{algorithm}

\subsection{Online Tuning of Reward Functions}

Upon establishing that the initial reward function is operational and can effectively achieve tasks in a simplified testing environment, the next step involves iteratively adjusting the reward function. However, judging the adequacy of the reward function by merely the magnitude of the rewards, as mentioned in \cite{eureka}, risks leading the LLM to amplify the reward coefficients or manipulate the reward structure, offering rewards for actions that do not authentically contribute to better completion of the task.

To this end, we have developed a high-level evaluation mechanism for the reward function, grounded in the task objectives. First, it is crucial to explain the purpose of the function. At each time step $t$, the reward function computes the rewards $r_t$ based on environmental observations. These rewards are then used alongside the predictions from the state value function $V(s_t)$ to calculate the Temporal Difference (TD) error: 
\begin{equation}
    \delta_t = r_t + \gamma V(s_{t-1}) - V(s_t).
\end{equation}
Subsequently, by aggregating the TD errors across $k$ time steps, we can derive :
\begin{equation}
\begin{aligned}
    \hat{A}_t^{(k)} &:= \sum_{l=0}^{k-1} \gamma^l \delta_{t+l}^V,
\end{aligned}
\end{equation}
Combining the aforementioned equation, the Generalized Advantage Estimator(GAE) \cite{GAE} can be defined as : 
\begin{equation}
\begin{aligned}
    \hat{A}_{t}^{\mathrm{GAE}(\gamma,\lambda)} 
    &= (1-\lambda) \sum_{l=0}^{\infty} \lambda^l \hat{A}_{t}^{(l+1)} \\
    &= (1-\lambda) \left( \delta_{t}^{V} + \lambda \sum_{l=0}^{\infty} (\gamma \lambda)^l \delta_{t+l}^{V} \right) \\
    &= \sum_{l=0}^{\infty} (\gamma \lambda)^l \delta_{t+l}^{V},
\end{aligned}
\end{equation}
where $\gamma$ is the discount factor that reduces the weight of future rewards, and $\lambda$ is another discount factor tunes the balance between immediate and long-term rewards. Here, we select $\gamma=0.99$ and $\lambda=0.95$.

Finally, the reward function can be defined according to the PPO algorithm \cite{PPO} as:
\begin{equation}
\begin{aligned}
\mathbf{loss}(\bm{\theta}^i) = \hat{\mathbb{E}}\Bigg[\min\Big(&r_t(\bm{\theta}^i) \hat{A}^{\text{GAE}(\gamma, \lambda)}_t, \\
&\mathrm{clip}\left(r_t(\bm{\theta}^i), 1-\epsilon, 1+\epsilon\right) \hat{A}^{\text{GAE}(\gamma, \lambda)}_t\Big)\Bigg],
\end{aligned}
\end{equation}
wherein, \(r_t(\bm{\theta^i}) = \frac{\pi(a_t, o_t; \bm{\theta^i})}{\pi(a_t, o_t; \bm{\theta^i_{\text{old}}})}\) represents the probability ratio between the new policy and the old policy. The term $\epsilon$ serves as a clipping factor to restrict large deviations between the new and old policies, here $\epsilon = 0.2$.
The loss function serves as a crucial indicator for policy convergence. Only when the loss function is trending towards convergence can we assert that the agents' actions are being guided by the current reward function. 
Subsequently, return the value of the following high-level evaluation metrics to the LLM:
\begin{itemize}
\item Success Rate: Reaching the destination without collisions.
\item Hazard Incidents: Instances where an agent comes too close to an obstacle.
\item Formation Error: Average deviation from the desired formation until reaching the destination.
\item Total Time: Duration taken to complete the task.
\item Average Acceleration: Mean absolute acceleration values until the destination is reached.
\end{itemize}
These metrics will assist in refining the reward function, ensuring it more accurately reflects the desired outcomes and improves the performance of the agents.

\begin{algorithm}[h]
\caption{Online Tuning of Reward Function}
\label{alg:llm-tune}
\KwIn{Initial reward $R_0$, policy $\pi_{\theta_i}$, value $\omega_i$, complex env. $\mathcal{E}$, iteration count $N$}
\KwOut{Updated policy $\pi_{\theta_i}$, value $\omega_i$}

\For{$k = 1$ to $N$}{
  Train $\pi_{\theta_i}^{(k)}, \omega_i^{(k)}$ using $R_{k-1}$ in $\mathcal{E}$ until loss converges\;
  Evaluate task metrics $\mathcal{M}_k$ (e.g., formation error, hazard, time)\;
  Update prompt $\mathcal{P}_k$ with $\mathcal{M}_k$ and policy summary\;
  Generate new reward $R_k \gets \texttt{LLM}(\mathcal{P}_k)$\;
}
\Return final $\pi_{\theta_i}, \omega_i$
\end{algorithm}

An objective-focused reward evaluation mechanism enables the LLM to more easily pinpoint adjustments needed for the reward function, resulting in enhanced task completion. The pseudocode is shown in Algorithm 2.


\section{EXPERIMENTAL RESULTS}
\label{EXPERIMENT}

\begin{figure}[!t]
    \centering
    \includegraphics[width=9cm]{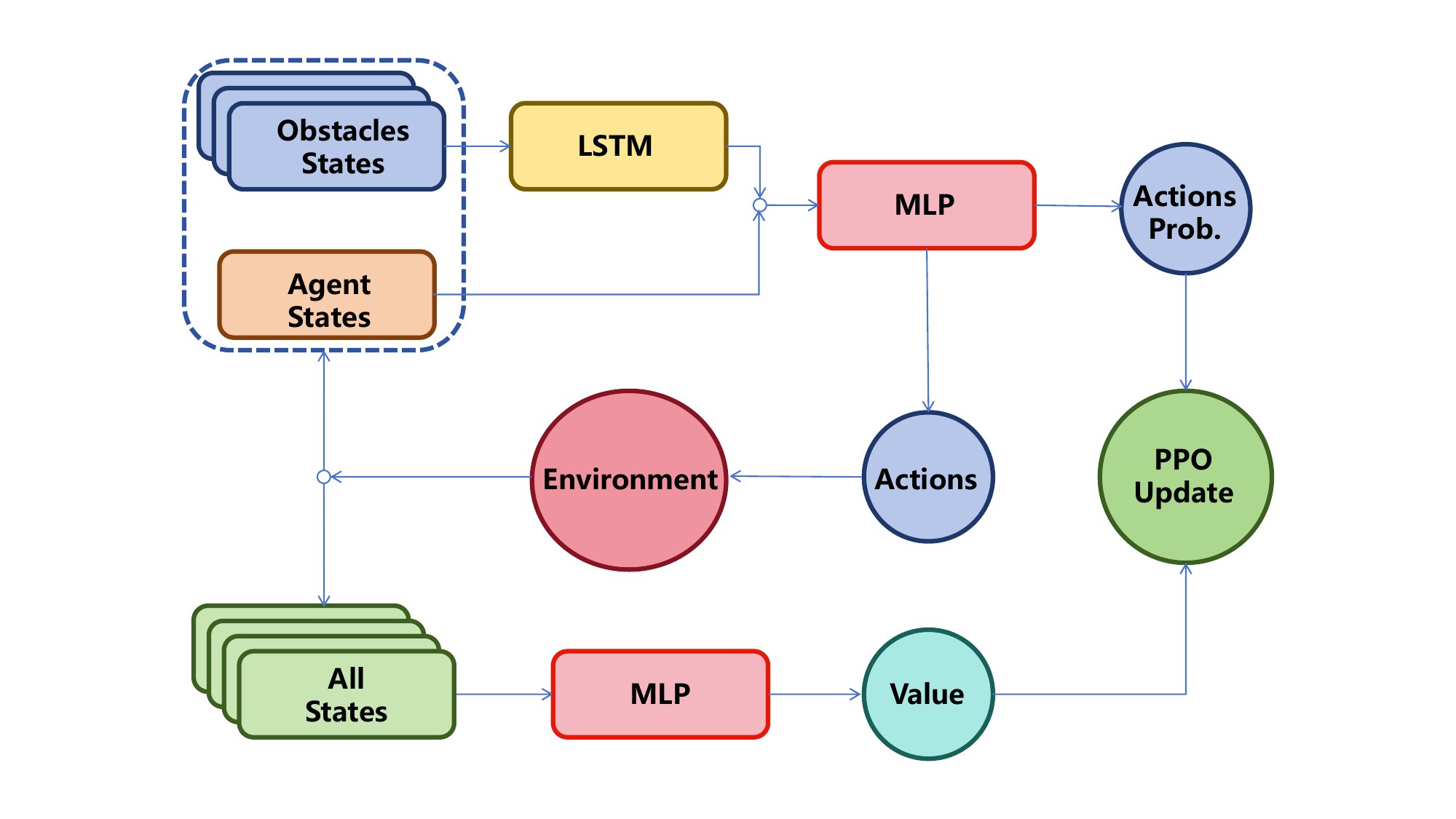}
    \caption{Training framework by using Actor-Critic method \cite{AC} and PPO algorithm}
    \label{training_framework}
\end{figure}

\subsection{Training Result}
We conducted the training in a custom simulation environment, with the specific training framework shown in Fig. \ref{training_framework}. 
In the policy network, the feature values of obstacles are extracted using both a Long Short Term Memory (LSTM) network with a hidden state size of 128 and a Multilayer Perceptron (MLP) with a hidden state size of 128. The feature values of the agent are extracted through a separate MLP with a hidden state size of 128. These extracted feature values (for both obstacles and the agent) are then concatenated and passed through a linear layer activated by ReLU to produce the final output. For the value network, all states are processed directly through an MLP with a hidden state size of 256, followed by a linear layer activated by ReLU to generate the output. Each episode involves 15 iterations of the PPO algorithm to refine and update the network weights.

We adopt a centralized training and decentralized execution (CTDE) framework. During training, data from all agents are aggregated, and a shared reward is used; however, each agent maintains its own individual policy network. During deployment, each agent makes decisions independently based on its own accessible observations and its respective policy network.

Qwen2.5-72B model \cite{Qwen2.5} is selected as the LLM to generate the reward function. Each iteration encompasses four key steps: generating the reward function, training, evaluation, and feeding the evaluation results back into the LLM to proceed to the next iteration. All evaluation results are summarized in \autoref{evaluation result}, the formation error in \autoref{evaluation result} is computed by \eqref{formation_error}. Each iteration consists of 20 episodes, with the reported results being the average performance across these episodes.

\begin{table}[h]
\caption{Evaluation results of all iterations}
\label{evaluation result}
\centering
\begin{tabular}{|c|c|c|c|}
    \hline
    \textbf{Iteration} & \textbf{Success rate (\%)} & \textbf{Average time} (s) & \textbf{Formation error}\\
    \hline
    0 & 60 & 11.0 & 74.6 \\
    \hline
    1 & 50 & 10.9 & 56.0 \\
    \hline
    2 & 95 & 11.5 & 73.8 \\
    \hline
    3 & 100 & 11.7 & 27.2 \\
    \hline
\end{tabular}
\end{table}

For the 0th iteration, both training and evaluation were conducted in a simplified environment featuring only three sparsely distributed dynamic obstacles. This setup aims to assess whether the LLM can comprehend the task and facilitate obtaining more substantial rewards in subsequent iterations. For all other iterations, the training and evaluation take place in a complex environment that includes seven densely distributed obstacles, combining static and dynamic elements. In this environment, both the obstacles and the agent have a maximum speed of 1.25 m/s.

Before proceeding to simulation and real-world deployment, it is important to clarify that the LLM is only involved in the reward function design during the training phase. Once the model training and evaluation are completed, the deployment and inference of the learned policy no longer rely on the LLM. In other words, the LLM does not affect the inference efficiency of the final trained model in either simulation or real-world deployment.


\begin{figure}[htbp]  
    \centering
    \subfloat[iteration0]{
        \includegraphics[width=0.95\linewidth]{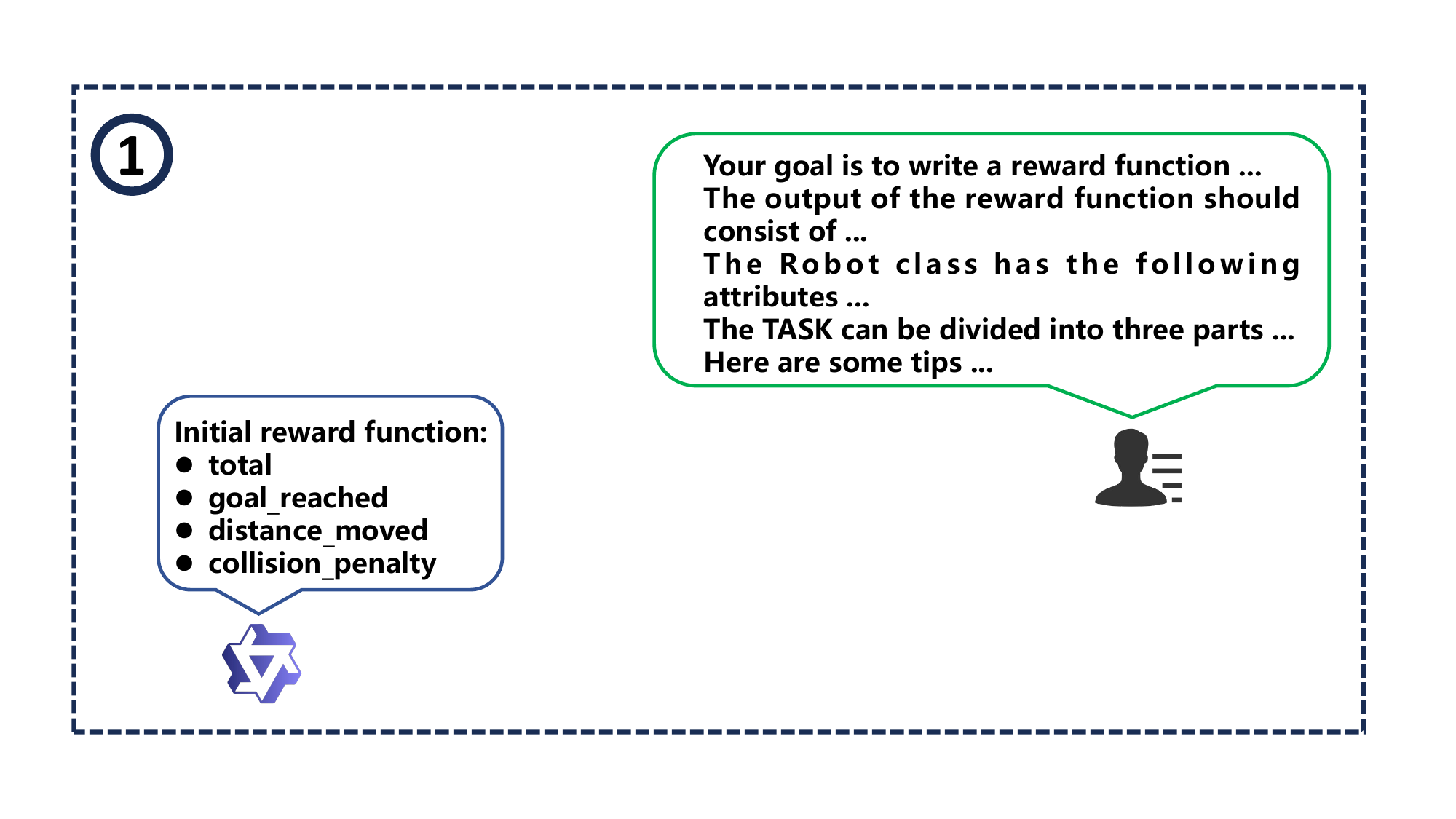}
        \label{fig:img1}
    }

    \subfloat[iteration1]{
        \includegraphics[width=0.95\linewidth]{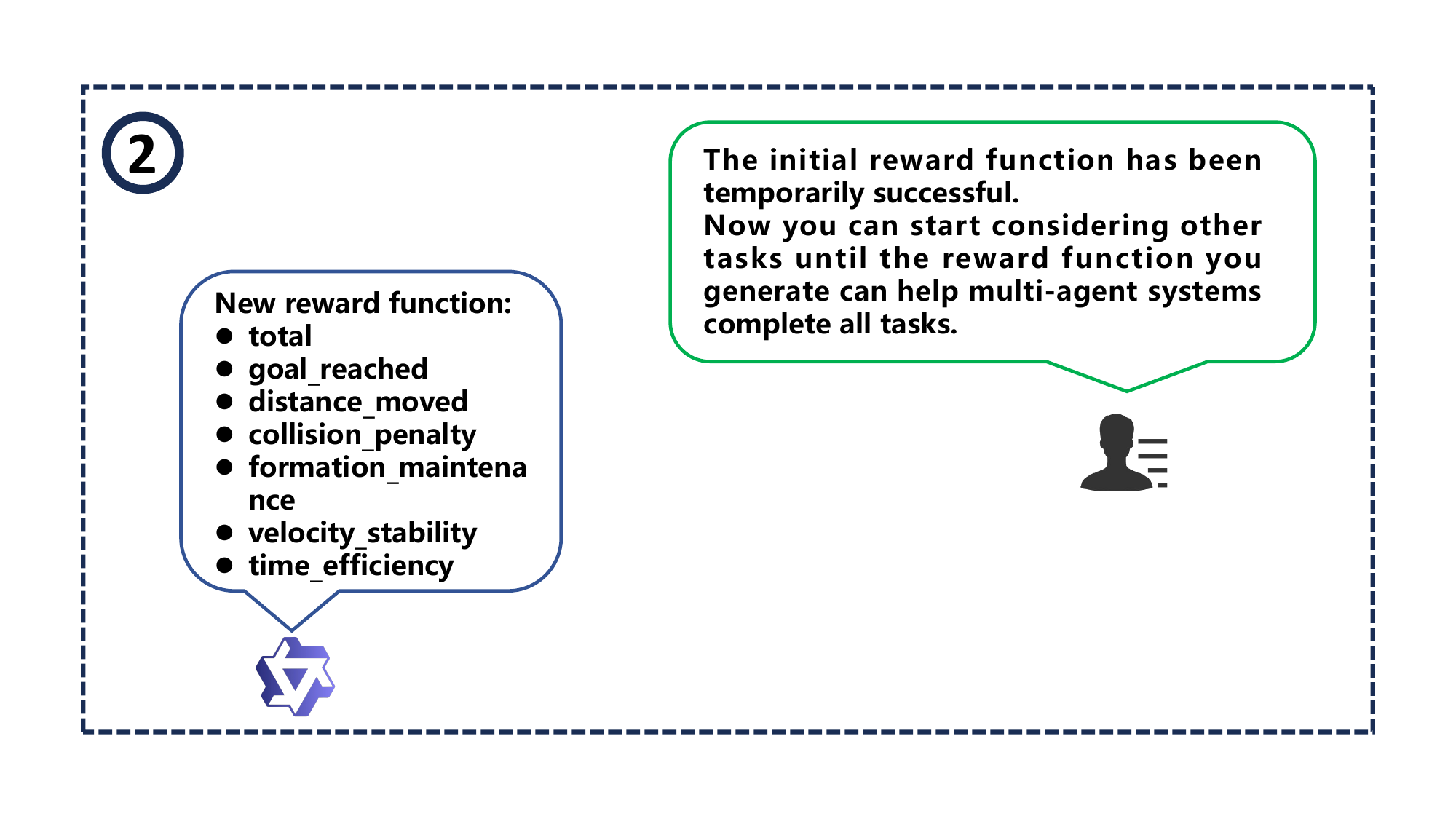}
        \label{fig:img2}
    }

    \subfloat[iteration2]{
        \includegraphics[width=0.95\linewidth]{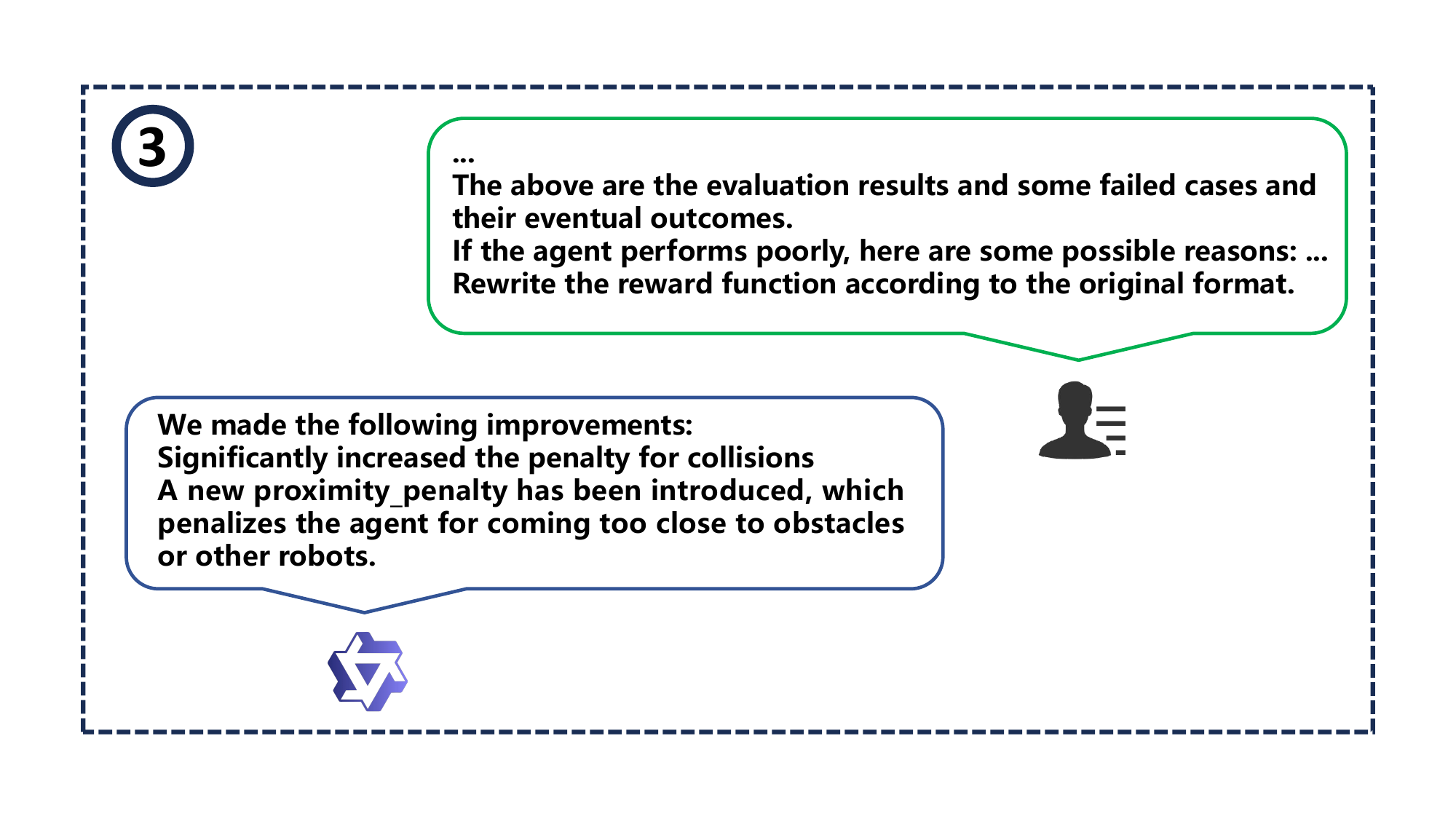}
        \label{fig:img3}
    }

    \subfloat[iteration3]{
        \includegraphics[width=0.95\linewidth]{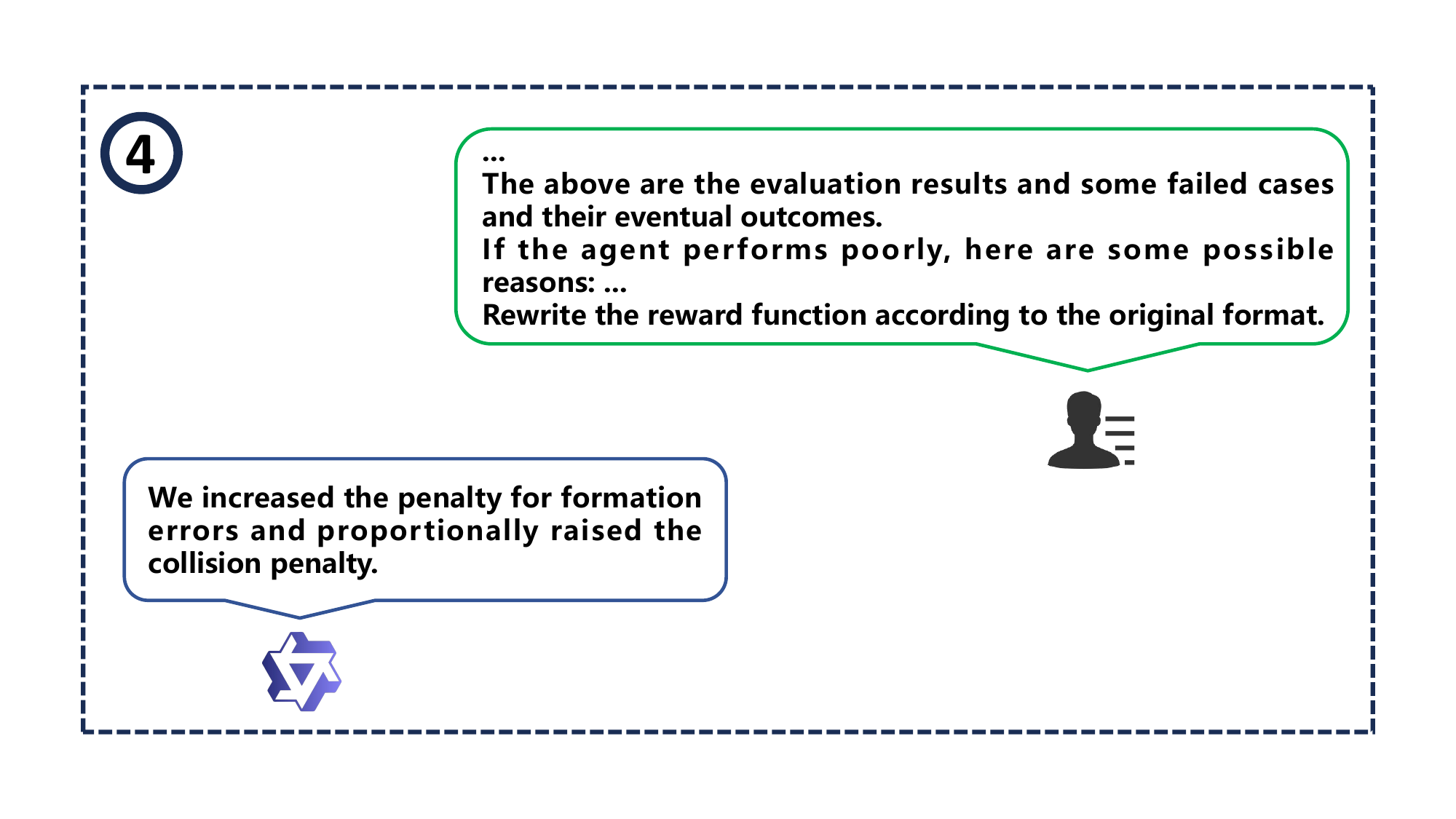}
        \label{fig:img4}
    }

    \caption{Summary of the conversation with LLM.}
    \label{LLM_dialogue}
\end{figure}

Fig. \ref{LLM_dialogue} summarizes the conversation content with the LLM. The initial reward function generated in the 0th iteration could only accomplish the tasks of reaching the destination and avoiding obstacles, without considering the interactions among various reward components. Starting from the 1st iteration, all tasks were considered; however, the results indicated that this iteration failed to properly balance these tasks, particularly performing poorly in obstacle avoidance. Feedback from this iteration was then provided back to the LLM, leading to improvements in the obstacle avoidance reward function during the 2nd iteration. Not only was the discount factor for the obstacle avoidance reward significantly increased, but a penalty for approaching obstacles was also introduced, substantially enhancing obstacle avoidance performance, though it overlooked the reward weight for formation maintenance. In the 3rd iteration, adjustments were made to increase the reward weight for formation maintenance while simultaneously adjusting the reward weights for obstacle avoidance and reaching the destination by a certain ratio. The final evaluation results demonstrated a 100\% success rate and showed excellent formation performance.

\begin{figure}[!t]
    \centering
    \includegraphics[width=9cm]{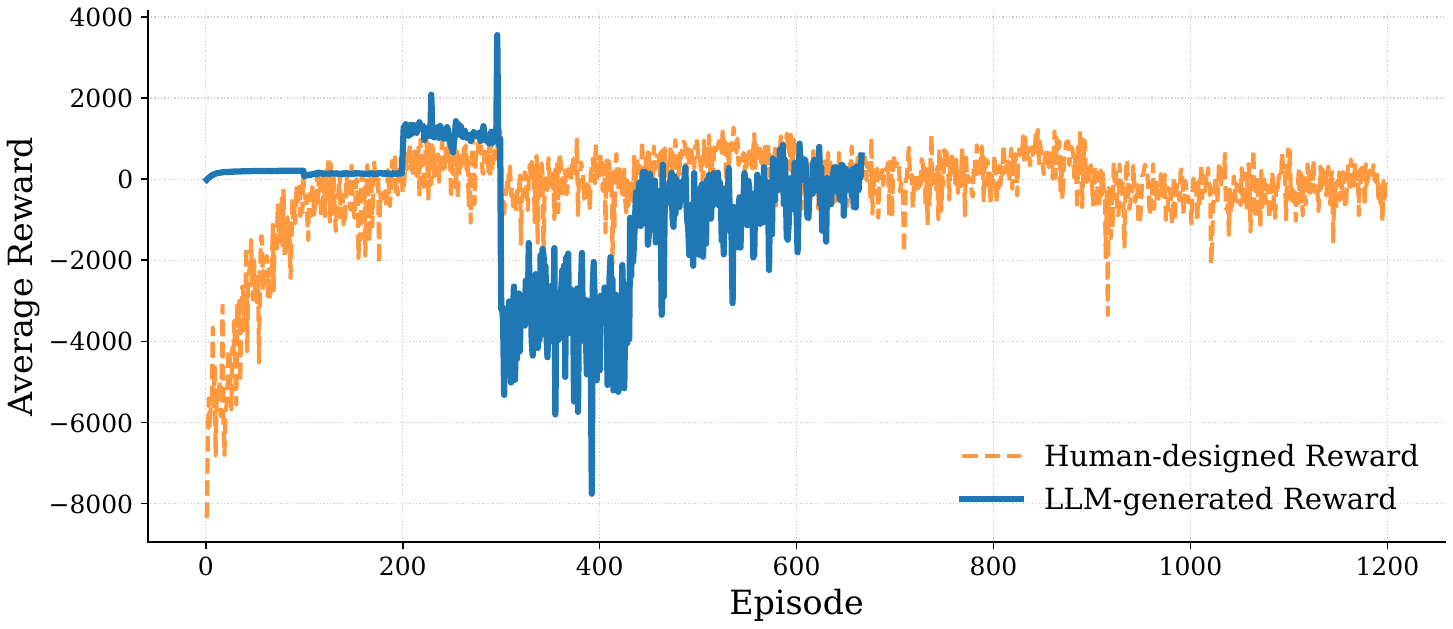}
    \caption{LLM-Generated vs Human-Designed Reward Functions}
    \label{reward_curve}
\end{figure}

To demonstrate the effectiveness of our approach, we compared it with a human-designed method \cite{mappo_rw}. Fig. \ref{reward_curve} illustrates the reward curves of the reward functions generated by LLM and those designed by humans during the training process, tracking the trend of reward changes from the start of training until the success rate reached 100\% during evaluation. The results show that the reward function generated and fine-tuned online by the LLM enabled agents to achieve a 100\% success rate more quickly in evaluations. 

Specifically, in the reward curves corresponding to the LLM-generated reward functions, the LLM redesigns the reward function in each iteration based on the evaluation of the previous version. This process can lead to substantial differences in the cumulative rewards achieved in the environment between two consecutive iterations. As a result, the reward curves may exhibit significant fluctuations across iterations—for example, a drastic increase in a penalty term may cause a sudden drop in the overall reward signal.

\begin{table}[h]
\caption{Comparison with other methods}
\label{tab:comparison}
\centering
\begin{tabular}{|c|c|c|c|}
    \hline
    \textbf{Method} & \multicolumn{1}{|p{1.5cm}|}{\centering Success\\rate (\%)} & \multicolumn{1}{|p{1.5cm}|}{\centering Average\\time (s)} & \multicolumn{1}{|p{1.5cm}|}{\centering Formation\\error} \\
    \hline
    ORCA-F & 79 & 19.2 & 35.2 \\
    \hline
    human-designed & 93 & 14.5 & 37.4 \\
    \hline
    LLM & \textbf{95} & \textbf{11.5} & \textbf{26.7} \\
    \hline
\end{tabular}
\end{table}

To further validate the universality of our model, we compared our approach with ORCA-F\cite{ORCA-F} and RL methods based on human-designed reward functions \cite{mappo_rw} by conducting evaluations, which use three different random seeds, each running for 300 episodes, and calculate the average performance across all runs. The results, as shown in \autoref{tab:comparison}, indicate that our approach outperforms the human-designed method in terms of success rate, time consumption, and formation error.

\subsection{Simulation and Real-World Deployment}

Given the uncertainty regarding the practical applicability of the model, direct deployment in real world could potentially result in damage. Therefore, we first validated the practicality of the model through simulations. To replicate real-world conditions as closely as possible within the simulation environment, both agents and obstacles were configured with physical parameters identical to those of their real-world counterparts. Considering that multi-agent systems rely on Robot Operating System (ROS) for communication, Gazebo, which offers the best integration with ROS, was selected as the simulation platform.

The simulated environment measures 20m $\times$ 20m, with seven obstacles initially positioned within a central 5m $\times$ 5m area. The agents are Mecanum wheel robots capable of omnidirectional movement, while dynamic obstacles are represented by TurtleBot2 robots, which use differential drive wheels. Both the diagonal lengths of the agents and the diameter of the obstacles are approximately 0.35m and both have a maximum speed of 1.25 m/s.

\begin{figure*}[t]
    \centering
    \includegraphics[width=18cm]{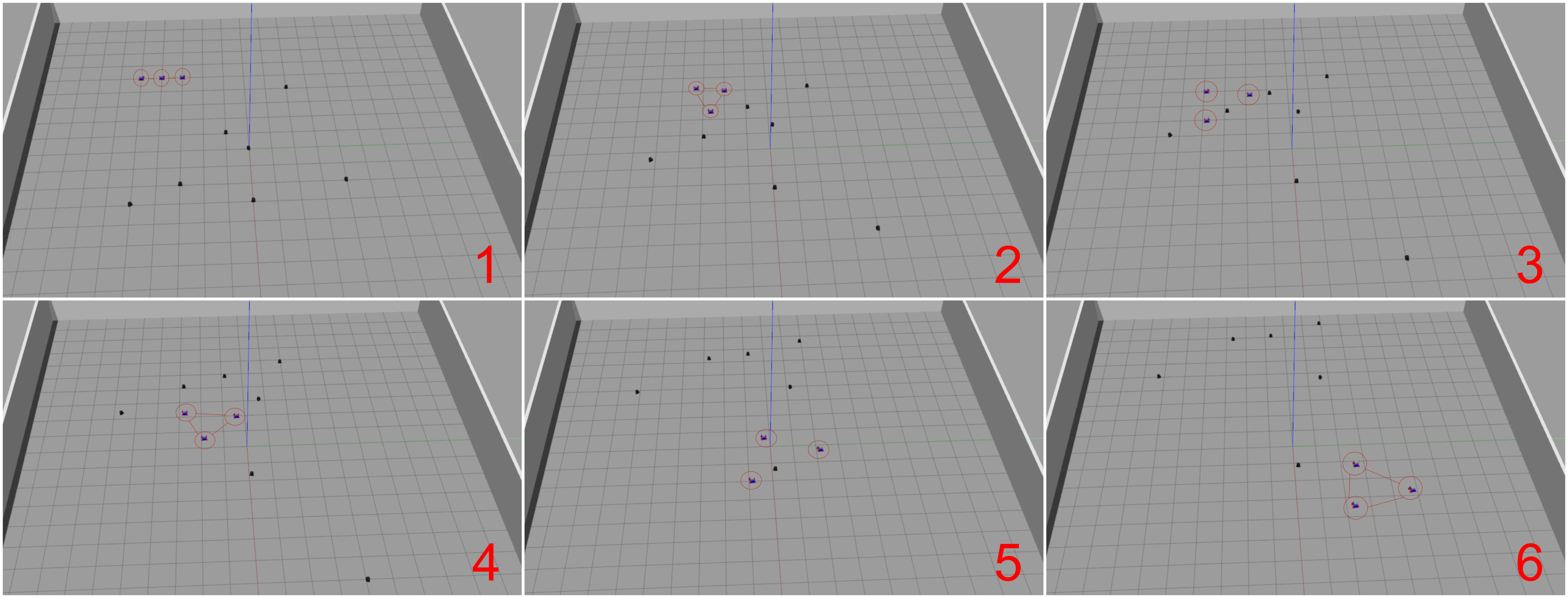}
    \caption{Snapshots of simulation}
    \label{simulation}
\end{figure*}

\begin{figure*}[!t]
    \centering
    \includegraphics[width=18cm]{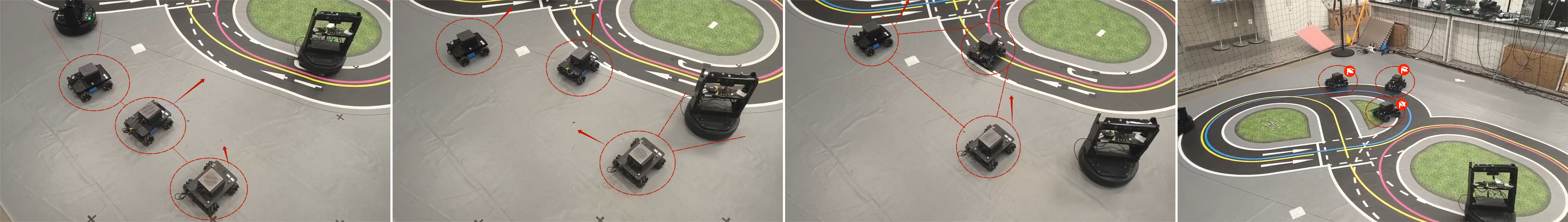}
    \caption{Snapshots of real-world deployment}
    \label{real-world}
\end{figure*}

Fig. \ref{simulation} presents snapshots of critical moments during the simulation. Initially, the three agents are aligned in a straight-line formation. Upon activation, they promptly reconfigure into an equilateral triangle formation. Throughout the simulation, the agents successfully maintain their formation while navigating around obstacles and swiftly re-establish the formation immediately after completing obstacle avoidance maneuvers.

Following successful validation in simulation, we conducted real-world experiments in a 5m $\times$ 5m environment. Each agent uses the OptiTrack motion capture system to obtain its own position and that of nearby obstacles, computing velocity via positional differentiation. Both agents and obstacles are limited to a maximum speed of 0.5m/s. During deployment, each agent runs on an NVIDIA Jetson AGX Orin, with decision-making at 10Hz—invoking the policy every 0.1s, though inference takes only 4ms. A low-level PD controller handles motor speed control at 240Hz.

Fig. \ref{real-world} showcases snapshots of critical moments from real-world experiments. The agents' performance mirrors that observed in the simulations: they successfully avoid all dynamic obstacles while maintaining an effective formation and reaching the destination. These experimental results demonstrate the efficacy of our approach in real-world applications.

\section{CONCLUSIONS}
\label{CONCLUSIONS}
This paper introduces a method for RL in FCCA that leverages LLM. By generating reward functions with LLM and establishing evaluation metrics for these functions, our approach facilitates more effective online tuning of RL reward functions. When applied to the FCCA problem, our method achieves success rates that are comparable to or even surpass those of human-designed reward functions. The efficacy and practicality of this approach have been confirmed through both simulation and real-world experimentation.

Despite its advantages, this method has notable limitations. For example, reward functions generated by LLM may lose track of initial requirements and early evaluation outcomes after several iterations. When tasks become more numerous and complex, LLM face significant challenges in balancing these tasks effectively.

In future work, we plan to deepen the integration of LLM with real-world multi-agent systems. By adopting methods analogous to those outlined in \cite{dreureka}, we aim to improve real-world performance and minimize the discrepancies between simulated and real-world environments.
 


\bibliography{Bibliography/IEEEabrv,Bibliography/IEEETrans}\ 

\end{document}